\documentclass[letterpaper]{article} 
\usepackage{aaai25}  
\usepackage{times}  
\usepackage{helvet}  
\usepackage{courier}  
\usepackage[hyphens]{url}  
\usepackage{graphicx} 
\usepackage{natbib}  
\usepackage{caption} 
\usepackage{algorithm}
\usepackage{algorithmic}

\usepackage{cite}
\usepackage{algorithmic}
\usepackage{graphicx}
\usepackage{textcomp}
\usepackage{bm}
\usepackage{tabularx,booktabs}
\usepackage{multirow}

\usepackage{algorithm}
\usepackage{amsmath}
\usepackage{adjustbox}
\usepackage{multirow}
\usepackage{amssymb}
\usepackage{anyfontsize}
\usepackage{blindtext}
\usepackage{multirow}
\usepackage{amssymb}
\usepackage{pifont}
\usepackage{booktabs} 
\usepackage{color, colortbl}
\usepackage[dvipsnames]{xcolor}
\usepackage[dvipsnames]{xcolor}
\usepackage[labelformat=simple]{subcaption}

\DeclareMathOperator*{\concat}{%
    \mathchoice%
        {\Big\Vert}%
        {\big\Vert}%
        {\Vert}%
        {\Vert}%
}

\usepackage{algorithm}
\usepackage{amsmath}
\usepackage{adjustbox}
\usepackage{multirow}
\usepackage{amssymb}
\usepackage{anyfontsize}
\usepackage{blindtext}
\usepackage{multirow}
\usepackage{amssymb}
\usepackage{pifont}
\usepackage{booktabs} 
\usepackage{color, colortbl}
\usepackage[dvipsnames]{xcolor}
\usepackage[dvipsnames]{xcolor}
\usepackage[labelformat=simple]{subcaption}

\usepackage{algorithm}
\usepackage{algorithmic}

\definecolor{Gray}{gray}{0.9}
\definecolor{LightCyan}{rgb}{0.88, 1, 1}

\usepackage{algorithm}
\usepackage{algorithmic}

\usepackage{newfloat}
\usepackage{listings}
\DeclareCaptionStyle{ruled}{labelfont=normalfont,labelsep=colon,strut=off} 
\floatstyle{ruled}
\newfloat{listing}{tb}{lst}{}
\floatname{listing}{Listing}
\title{Halal or Not: Knowledge Graph Completion for Predicting Cultural Appropriateness of Daily Products}

\author{
    Van Thuy Hoang$^{1}$, 
    Tien-Bach-Thanh Do$^{1}$,
    Jinho Seo$^2$,
    Seung Charlie Kim$^3$,
    Luong Vuong Nguyen$^4$,
    Duong Nguyen Minh Huy$^4$,
    Hyeon-Ju Jeon$^{5,*}$, 
    and O-Joun Lee$^{1,*}$
}
\affiliations{

    $^1$Department of Artificial Intelligence, The Catholic University of Korea, Republic of Korea \\
    $^2${School of Computer Science and Information Engineering, The Catholic University of Korea, Republic of Korea}\\
    $^3$Shukran Korea Co., Ltd., Republic of Korea \\
    $^4$FPT University Danang Campus, Vietnam \\
    $^5$Data Assimilation Group, Korea Institute of Atmospheric Prediction Systems (KIAPS), Republic of Korea \\
    {\{hoangvanthuy90, osfa19730, ojlee\}@catholic.ac.kr}, 
    kkinho04@catholic.ac.kr, 
    sck@shukrankorea.com, 
    {\{vuongnl3, huydnm3\}@fe.edu.vn}, 
    hjjeon@kiaps.org
    
}

\usepackage{bibentry}

\begin{document}

\maketitle





\begin{abstract}

The growing demand for halal cosmetic products has exposed significant challenges, especially in Muslim-majority countries.
Recently, various machine learning-based strategies, e.g., image-based methods, have shown remarkable success in predicting the halal status of cosmetics.
However, these methods mainly focus on analyzing the discrete and specific ingredients within separate cosmetics, which ignore the high-order and complex relations between cosmetics and ingredients.
To address this problem, we propose a halal cosmetic recommendation framework, namely HaCKG, that leverages a knowledge graph of cosmetics and their ingredients to explicitly model and capture the relationships between cosmetics and their components.
By representing cosmetics and ingredients as entities within the knowledge graph, HaCKG effectively learns the high-order and complex relations between entities, offering a robust method for predicting halal status.
Specifically, we first construct a cosmetic knowledge graph representing the relations between various cosmetics, ingredients, and their properties. 
We then propose a pre-trained relational graph attention network model with residual connections to learn the structural relation between entities in the knowledge graph.
The pre-trained model is then fine-tuned on downstream cosmetic data to predict halal status.
Extensive experiments on the cosmetic dataset over halal prediction tasks demonstrate the superiority of our model over state-of-the-art baselines.

\end{abstract}



\maketitle

\section{Introduction}
\label{sec:introduction}

The cosmetic industry includes various products and ingredients, ranging from skincare and perfumes to other items designed for beauty enhancement and personal care \cite{rauda2022convolutional}. 
These cosmetic products often feature a variety of natural, synthetic, and innovative ingredients designed to cater to consumers' evolving needs and preferences.
However, awareness of cosmetics' ingredients does not always align with consumer consumption patterns, particularly when it comes to distinguishing between halal and haram cosmetics. 
For instance, most consumers often neglect to check labels or review the detailed composition of beauty products.
This oversight may lead to the ongoing use of products without fully understanding their ingredients or possible effects \cite{rauda2022convolutional,kim2011porcine}.
Furthermore, although awareness of halal standards and ingredients in cosmetic products is growing among consumers, identifying halal versus non-halal products remains challenging. 
These challenges arise due to the diverse range of brands, each offering products with different ingredient compositions and chemical formulations \cite{park2007antioxidative}. 


Halal ingredients are necessary for Muslims to comply with dietary laws outlined in Islamic principles.
For example, the ingredients within products should not include prohibited things like pork, alcohol, or any animal, according to Islamic guidelines \cite{rauda2022convolutional}.
With the growing demand for halal products globally, many industries, from food to drugs, are working to ensure their products comply with halal standards.
The growing demand for halal products has led to the establishment of numerous halal certification institutes worldwide.
Currently, there are more than 500 halal certification institutions worldwide \cite{tieman2019creative}.
However, no international board manages a centralized database to unify halal certification data across institutions \cite{rakhmawati2021linked}. 
Consequently, a single product may carry multiple halal certificates from various organizations since each institution has its own standards and regulations. 

Recently, machine learning-based methods have been applied to identify halal ingredients in cosmetic products \cite{karimah2021effect,cetin2016muslim}.
The adoption of machine learning-based methods for halal certification introduces both opportunities and challenges that have significant ethical and societal implications.
These methods offer transformative potential in addressing the challenges of halal certification and making halal certification more accessible, especially for small organizations that struggle to meet traditional methods' cost or time requirements.

Inspired by the success of the image processing methods in computer vision, the halal status of products can potentially be identified via image processing methods, e.g., Optical Character Recognition (OCR), that analyze the ingredient lists and product labels \cite{karimah2021effect}.
One of the most famous rules is the absence of animal or pork ingredients \cite{cetin2016muslim,jia2021turning}.
Through image processing, pattern recognition models can be applied to detect specific text patterns related to ingredients on individual packaging \cite{gultekin2020evaluation}. 
Although the strategies can assist in verifying halal compliance for ingredients, interpreting product labels and ingredient sourcing remains a significant challenge.
That is, ingredients are often listed under scientific or chemical expressions, making it difficult for image-based strategies to identify halal or non-halal components.
Furthermore, these methods face significant technical challenges when dealing with suboptimal conditions during text extraction.
For example, pixelated images in poor lighting conditions can hinder accurate text recognition.
To sum up, most existing studies mainly focus on analyzing the discrete and specific cosmetics, resulting in ignoring capturing the high-order and complex relationships between cosmetics and ingredients.



To overcome the limitations, we introduce HaCKG, a recommendation framework that leverages knowledge graph representation learning to predict halal standards. 
A knowledge graph, often called a semantic network, models a network of real-world entities and captures their relationships \cite{lin_2020_kgnn,wang_2019}. 
In the context of cosmetic ingredients, the knowledge graph can represent the relationships between products, ingredients, and their associated attributes.
By representing the relations of such cosmetic products, we could then explore their interactions and complex relations through graph-based machine learning algorithms \cite{xu2018powerful}.
Furthermore, as cosmetics and the ingredients are represented in our knowledge graph, the sparse cosmetic data can be enriched with additional context, e.g., their ingredients and properties relations, which could help our model capture all contexts for cosmetics and ingredients.
Specifically, we extract eleven entity types and five relation types from the cosmetics and ingredients to build our knowledge graph. 
Then, as the ingredient properties contain numerical attributes, we design a fusion layer incorporating a gate function to capture different attribute types effectively into unified input features.
We then propose a relational Graph Attention Network (r-GAT) with a residual connection to learn the cosmetic representations. 
The r-GAT model is pre-trained in a Self-Supervised Learning (SSL) manner without using any label information. 
Then, the pre-trained model can be fine-tuned to determine the likelihood that a given product meets halal standards. 
To the extent of our knowledge, we are the first to represent cosmetic products in knowledge graphs and learn their relations through graph neural networks.

In summary, our contributions are as follows:
\begin{itemize}
    
    \item We construct a cosmetic knowledge graph that represents the natural relations between cosmetic products, ingredients, and the ingredient’s properties. 
    The knowledge graph then can serve as a fundamental tool for our model to learn cosmetic relations and predict the halal status.
    \item We propose a pre-trained residual Graph Attention Network (r-GAT) that incorporates residual connections to capture relationships between entities in the cosmetic knowledge graph effectively. 
    The pre-training strategy enables r-GAT to learn the structural relationships among cosmetics, ingredients, and their properties without relying on labeled data, allowing for efficient fine-tuning on specific cosmetic product tasks.
    
    
    \item  We conduct extensive experiments on halal cosmetic product datasets. 
    The significant improvements demonstrate the superiority of our proposed model compared to state-of-the-art baselines.
    
\end{itemize}

\section{Related work}

We now discuss several machine learning-based strategies to identify the status of cosmetic products compared to our strategy.
The recent strategies can be categorized into three main approaches: text-based strategies, image processing-based strategies, and graph-based strategies.

Text-based strategies focus on analyzing ingredient lists or product descriptions to determine compliance using a large text corpus.
For example, DIETHUB \cite{petkovic2021diethub} is a tool designed to predict and recommend product recipes based on a hierarchy, describing food entity relations from Hansard's corpus \cite{mollin2007hansard} with various food entity annotations and recipes.
The large number of product entities in the corpus enables various ingredients and recipes, catering to different dietary preferences and restrictions, including halal.
They then use a language model, e.g., doc2vec, to learn the hierarchy relations between ingredients, products, and their labels \cite{vrehuuvrek2010software}.
Ispirova et al. \cite{ispirova2020p} use a set of products representing various items from Slovenian food consumption data. 
They then combine word embedding and graph learning to cluster products into separate categories. 

For image processing-based strategies, several methods adopt convolutional neural networks (CNNs) to recognize halal products and their components.
For example, Ramdania \cite{rauda2022convolutional} introduced a convolutional neural network to recognize letters or characters in the ingredients.
Other strategies utilize CNNs and OCR tools to extract text from images and recognize the ingredient letters from images. 
CNNs can be used with OCR tools, e.g., Tesseract \cite{smith2007overview}, to extract text from images, which can be integrated to convert image data into text.
However, only checking the presence of ingredients can be insufficient to decide whether the products are halal or non-halal. 
This is because alternatively sourced ingredients may still be allowed to be part of a halal cosmetic product.
For example, cosmetic products may contain ethanol as long as it is sourced from natural aerobic fermentation (e.g., natural fermentation process in the presence of oxygen) or synthetic sources (i.e., prepared from ethylene oxide, acetaldehyde, acetylene) and not from the liquor industry.
Moreover, by considering individual products, the existing methods fail to learn the complex relations between many entities, such as ingredients and products. 
That is, these methods could ignore the high-order connectivity, which aligns with the nature of the relationships between cosmetics and ingredients.

Several graph-based studies have been proposed to learn the cosmetic representations and ingredients by optimizing their similarity \cite{DBLP:journals/mis/LeeHK21,nguyen2023connector,s23084168,hoang2023mitigating}.
For example, Rakhmawati et al. \cite {rakhmawati2022halal} collected Halal product datasets and then explored their relations with several simple similarity tools, such as Jaccard or nearest neighbor similarity.
Rakhmawati and Najib \cite{rakhmawati2021linked} adopted Node2Vec to learn the similarity between products and ingredients. 
By constructing a knowledge graph, Rakhmawati et al. \cite{rakhmawati2023halal} capture the relations between cosmetics based on shared ingredients.
They then use several basis graph algorithms to build entity features, such as common neighbor and label propagation, to put into a basis machine learning models, e.g., random forest and k-nearest neighbors algorithm.
While such models can easily apply such similarities and distances, the model could not learn heterogeneous properties and complex relations between ingredients and products \cite{DBLP:journals/corr/abs-2308-09517}.
Several knowledge graph-based methods extend traditional GNNs to capture higher-order relationships within the knowledge graph \cite{lin_2020_kgnn,wang_2019,Lee2020a,DBLP:journals/sensors/JeonCL22,DBLP:journals/joi/LeeJJ21}.
For example, KGAT \cite{wang_2019} applies an attention mechanism to weigh the influence of each neighboring entity differently across multiple hops in the recommendation knowledge graph.
LiteralKG \cite{DBLP:journals/access/HoangNLLNL23} introduces an attentive propagation with residual connection and identity mapping to capture the complex relations between entities in the medical knowledge graph.
In contrast, by representing such relations between cosmetics in the knowledge graph, our model could learn the representations and capture the relationship between various entities based on the relational attention mechanism.



\section{Methodology}

\begin{figure*}[t]
\centering 
  \includegraphics[width= 1 \linewidth]{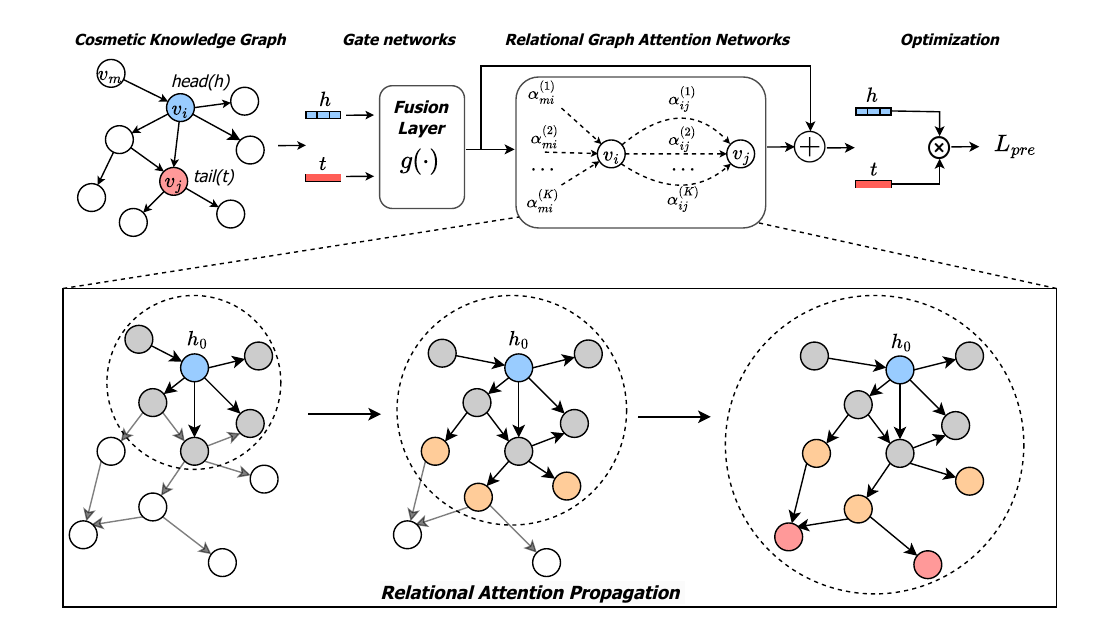}
  \caption{The overall architecture of HaCKG.
  The model comprises four main blocks: cosmetic knowledge graph construction, gate networks, relational Graph Attention Networks, and Optimization.
  }  
  \label{fig:model}
\end{figure*} 

In this section, we first explain how the cosmetic knowledge graph can be constructed from cosmetics, ingredients, and their properties.
Then, we present the model architecture, including fusion layers and attentive propagation with residual connection.
Lastly, we introduce our strategy for the pre-training phase without using any label information about cosmetic status, followed by the fine-tuning phase.
Figure \ref{fig:model} shows an overview architecture of HaCKG.

\begin{figure*}[t]
\centering 
  \includegraphics[width= 1 \linewidth]{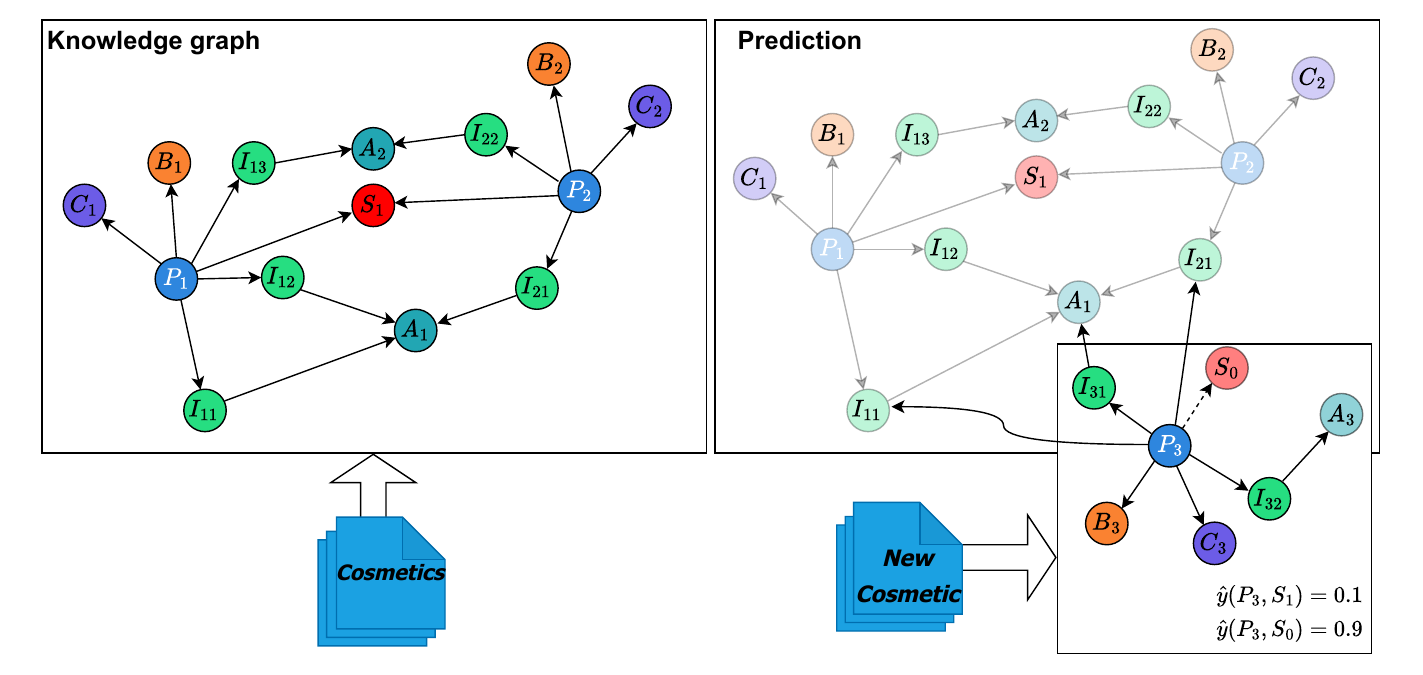}
  \caption{An example of the product prediction given a knowledge graph.
  The pre-trained model can predict a score between a new product $P_3$ and the status $S_0$ or $S_1$.}  
  \label{fig:predict}
\end{figure*}

\subsection{Cosmetic Knowledge Graph Construction}

We now represent our strategy to construct a cosmetic knowledge graph from cosmetic product records.
A knowledge graph (KG) is a semantic framework that models heterogeneous data by connecting entities and their relationships, effectively mirroring real-world relationships \cite{bordes_2013,lin_2015}. 
Formally, a KG is a set of triples where each triple is formed of $\langle h, r, t \rangle$, where $h$, $r$, and $t$ refer to the head, relation, and tail, respectively \cite{dai_2020_survey}.
That is, each triplet $\langle h_j, r_j, t_j \rangle$ contains pairs of entities $\langle h_j \ , t_j \rangle$ and a corresponding relation type $r_j$.
A cosmetic knowledge graph (CKG) is a knowledge graph that represents the relations between cosmetic products, ingredients, and their properties \cite{gong_2021}.

We now explain our strategy to construct a cosmetic knowledge graph and entity relations from cosmetic product records.
Given a cosmetic product dataset collected from Shukran Korea Co., Ltd.\footnote{\url{https://www.shukrankorea.com/}}, each record contains a set of information properties, such as product brands, product names, ingredients, and their properties.
We then transform the objects and properties from the records into entities and build our cosmetic knowledge graph.
In our cosmetic knowledge graph, entities can be cosmetic products, ingredients, and properties.
For example, in each record, we extract the entity names from the product name fields.
Table~\ref{tab:entity_types} shows the detailed statistics of entity types.
Our cosmetic knowledge graph has eleven entity types, including cosmetics, ingredients, product brands, categories, cosmetic status, and ingredient properties.
Table \ref{tab:relation_type} represents the five relation types between entities.
That is, we construct different types of edges that connect cosmetics, ingredients, brands, and ingredient properties.
For example, $r_i$ refers to the relation between cosmetics and its ingredients, while $r_b$ presents the connection between cosmetics and brands.
As a result, we constructed 101,186 entities with different entity types and relations in our cosmetic knowledge graph.
We believe that by representing the products, ingredients, and properties through the knowledge graph, we could capture the natural relationships between cosmetic products, which could then benefit the model in learning cosmetic relations.


\begin{table*}[t]
\centering
\caption{A summary of entity types in our cosmetic knowledge graph.}
\label{tab:entity_types}
\begin{tabular}{p{0.5cm} p{3 cm} p{1.2 cm} p{6 cm} }
\toprule
\textbf{\#} & \textbf{Entity type} & \textbf{Notation} &  \textbf{Description}   \\\midrule
1&	Cosmetic &	P	&The name of cosmetic	 \\
2&	Ingredient &	I	&The name of the ingredient 	 \\
3&	Brand &	B	& The cosmetic brand	 \\
4&	Category &	C	&The cosmetic category	 \\
5&	Ingredient property 1 &	T	&Toxicity state of the ingredients \\
6&	Ingredient property 2 &	A	&Allergy state of the ingredients \\
7&	Ingredient property 3 &	Ca	&Cancer state of the ingredients	 \\
8&	Ingredient property 4 &	R	&Restriction state of the ingredients	 \\
9&	Ingredient property 5 &	$S_{mi}$	&The minimum score of the ingredients	 \\
10&	Ingredient property 6 &	$S_{ma}$	&The maximum score of the ingredients	 \\
11&	Status& S	&The Halal/Haram status of the cosmetic \\
\bottomrule
\end{tabular}
\label{tab:KG}
\end{table*}
 
\begin{table}[t]
\centering
\caption{A summary of relation types in our cosmetic knowledge graph.}
\label{tab:datasets}

\begin{tabular}{ p{0.2 cm}  p{0.6 cm}  p{6 cm} }
\toprule
\textbf{\#} & \textbf{Type}  &  \textbf{Description}   \\\midrule
1&	$r_{i}$&The relation between cosmetics and its ingredients \\
2&	$r_{b}$&The relation between cosmetics and their brand  \\
3&	$r_{c}$&The relation between cosmetics and their category  \\
4&	$r_{p}$&The relation between ingredients and their properties	 \\
5&	$r_{s}$&The relation between cosmetics and its status \\
\bottomrule
\end{tabular}
\label{tab:relation_type}
\end{table}

\begin{figure}[t]
\centering 
\includegraphics[width= \linewidth]{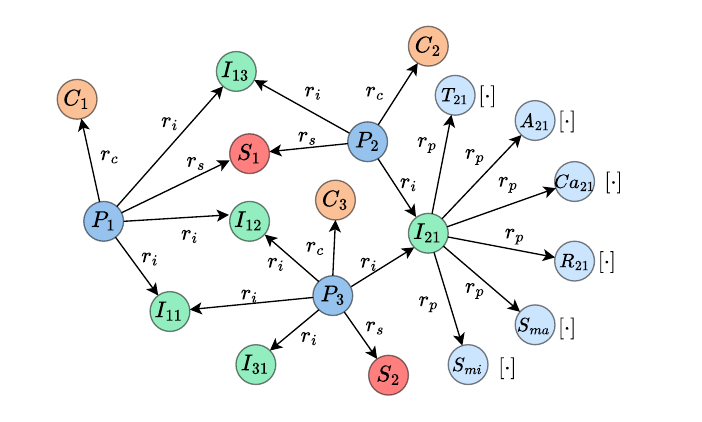}
\caption{An example of the cosmetic knowledge graph.
The circles and narrows denote entities and their relations, respectively. 
There are several entities, such as $P$,  $B$, and $I$, denote the cosmetic product, cosmetic brand, and ingredients, respectively.
The brackets $[\cdot]$ refer to the numeric attributes of the ingredient properties.}
\label{fig:KG_construction}
\end{figure}

To illustrate our cosmetic knowledge graph construction, Figure \ref{fig:KG_construction} shows three cosmetic products ($P_1$, $P_2$, and $P_3$), their ingredients, properties, and their relations.
Let $P$ denote the product entities, and $I$ refer to the set of product ingredients for each cosmetic product.
We construct one-to-many relations between product entity $P$ and $I$ as a cosmetic product $P_i$ can contain many ingredients.
Additionally, other entities within our CKG are connected one-to-one with their corresponding product entities, reflecting the inherent nature of their relationships.
Other entities in our CKG are constructed and connected to their product entities $P$ as a one-to-one connection that satisfies their natural types of relations. 
For example, the product $P_1$ is a cosmetic product from a brand $B_1$ and has ingredient $I_{11}$ as its properties.
Moreover, the ingredients also can have many properties, such as Toxic properties ($T$) and Allergy ($A$), as shown in Figure \ref{fig:KG_construction}.



\subsection{Fusing Attributes and Entities Features}

As mentioned earlier, several ingredient properties, e.g., toxicity or allergy, could contain numeric information, which delivers the state of the ingredient.
Therefore, we first design a fusion layer that contains a gate function to transform different types of attributes and entities into unified representations as initial inputs for our model.
That is, numerical attributes, such as ingredient properties, are normalized and transformed directly into feature vectors \cite{DBLP:conf/semweb/KristiadiKL0F19}.
We then transform entities into the shared space to generate the unified numerical-enriched vectors using a gate function \cite{kristiadi_2019}. 
Specifically, for each entity $i$, we first transform the initial entity vector $h_i \in R^{1 \times E}$ and numerical vector ${n}_{i}\in R^{1 \times N}$ into ${{{\hat{h}}}_{i}}$ and ${{{\hat{n}}}_{i}}$, as:
\begin{align}
& {{{\hat{h}}}_{i}} = W_E \cdot h_i, \\ 
& {{{\hat{n}}}_{i}} = W_N \cdot n_i,
\end{align}
where $W_E \in R^{E \times d}$ and $W_N \in R^{N \times d}$ are learnable parameters, $E$ and $N$ refer to the dimensions of initial entities and numerical vectors.
The entity vector ${{{\hat{h}}}_{i}}$ and numerical attribute vector ${{{\hat{n}}}_{i}}$ are then combined and transformed into a vector ${{e}_{i}}$, as:
\begin{align}
& {{e}_{i}}=g\left( {{h}_{i}},{{n}_{i}} \right)=\alpha \odot \beta +\left( 1-\alpha  \right)\odot {{{\hat{h}}}_{i}}, \\ 
&\text{where} \ \alpha ={{\sigma }_{1}}\left( {{{\hat{h}}}_{i}}+{{{\hat{n}}}_{i}} \right),  \notag \\
& \ \ \ \ \  \ \ \ \ \ \text{ }\beta ={{\sigma }_{2}}\left( W\left( {{h}_{i}}\lVert{{n}_{i}} \right) \right),  \notag
\end{align}
where $h_i$ and $n_i$ refer to the initial features of entities and numeric entities, respectively.
$\odot$ is the Hadamard product, and $\sigma_1 (\cdot)$ and $\sigma_2(\cdot)$ are $\text{Sigmoid}(\cdot)$ and $\text{Tanh}(\cdot)$ activation functions, respectively.
$W \in R^{(E+N) \times d}$ is the learnable parameter.

\subsection{Learning Relational Attention}

To account for the diverse types of relationships between nodes in our cosmetic knowledge graph, we propose employing a relational Graph Attention Network (r-GAT) with residual connection \cite{DBLP:journals/corr/abs-2109-05922,chen_2020}. 
Specifically, to capture the importance of a neighboring entity to a given entity, we calculate the attention coefficients for measuring the significance of a neighbor entity $u$ for a given entity $v$.
These coefficients are then normalized using a softmax function, guaranteeing that the attention scores for all neighboring nodes of a given node sum to one.
The attention coefficient of each entity $u$ contributes to the given entity $v$ at layer $l$-th on channel $k$ as follows:
\begin{align}
\alpha _{viu}^{k}=\frac{\exp \left( f^k \left(e_{v}^{k}\lVert r_{i}^{k}\lVert e_{u}^{k} \right) \right)}{\sum\limits_{z\in {{N}_{v}}} \sum\limits_{j\in {R_{zv}}}{ {\exp \left( f^k \left( e_{v}^{k}\lVert r_{j}^{k}\lVert e_{z}^{k} \right) \right)}}},
\end{align}
where $N_v$ is the set of the neighbors of entity $v$,
$f^k$ is a feedforward neural network, 
$r_i$ refers to the relation type $i$ between entity $u$ and $v$. 
$R_{vz}$ denotes the relations between entity $v$ and the neighbour $z$. 
$r_{j}^{k}$ denotes the relation between entity $v$ and the neighbour $z$. 
To capture the attention of neighbor $u$ to $v$ at the $k$-th channel, we combine its embeddings and relation types.
Specifically, the embeddings of entity $v$ at the $l$-th layer can be computed as:
\begin{align}
e_{v}^{k(l)}=\sigma \left( \sum\limits_{u\in {{N}_{v}}} \sum\limits_{i\in {R_{vu}}} { {\alpha _{viu}^{k}\left[ e_{u}^{k}*r_{i}^{k} \right]}} \right),
\end{align}
Finally, the representation of the entity $v$ is obtained by a concatenation operator as:
\begin{equation}
e_{v}^{(l)}=\underset{k=1}{\overset{K}{\concat}}\,\sigma \left( \sum\limits_{u\in {{N}_{v}}} \sum\limits_{i\in {R_{vu}}} {{\alpha _{viu}^{k}\left[ e_{u}^{k}*r_{i}^{k} \right]}} \right),
\end{equation}
where $\lVert $ denotes the concatenation. 
However, stacking multiple GNN layers can lead to the over-smoothing problem, where the representations of nodes converge and become nearly identical \cite{chen_2020}. 
Thus, we add initial residual connections to the entity representations at each GNN layer. 
Formally, the representations of the entity $v$ can be computed as:
\begin{align}
& e_{v}^{(l+1)}=\sigma \left( \left( \left( 1-{{\alpha }^{(l)}} \right) {P} e_{v}^{(l)} \right)+{{\alpha }^{(l)}}e_{v}^{(0)} \right), \
\end{align}
where ${P}$ refers to a graph convolution matrix, i.e., ${P}={{ \tilde{D}}^{-1/2}} \tilde{A}  {\tilde{D} ^{-1/2}}$, $\alpha$ is the hyper-parameter, and $\tilde{D} = D + I $ and $\tilde{A} = A+I$ are the degree and adjacency matrix, respectively, $I$ is the identity matrix, and $\alpha$ is a hyperparameter.

To compute the final representation of an $i$-th entity $e_{i}$, we first concatenate all the output representations of GNN layers. 
As the local and global graph structures are essential to represent entities, we aim to combine the output features of all GNN layers.
Therefore, the representations could learn the local and global graph structures \cite{wang_2019, hong_2022_lagat}.
We then apply a linear function followed by an activation function to transform the entity vectors into final representations as:

\begin{equation}
\label{eq11}
e_i = \sigma \left(W \cdot {\concat}_{l=1}^L \left(e_{i}^{(l)}\right) + b\right),
\end{equation}
where $L$ presents the number of GNN layers, ${b}$ is a learnable parameter, and $W$ is the trainable weight matrix.

\subsection{Model Optimization}

We now introduce our pre-training strategy for HaCKG, which aims to preserve the co-coefficients across all triplets through a scoring function without using any label information.
That is, the model will learn entity representations to preserve the graph structure and relations between entities.
To preserve the entity relations, we aim to maximize all the positive triplets from our Cosmetic Knowledge Graph and minimize negative triplets that do not come from KGs \cite{bordes_2013,lin_2015}.
Specifically, for each triplet, entity embedding vectors are first transformed into a shared relation space through a projection matrix, followed by a scoring function to compute the coefficients between entities.
The scoring function can be defined as:
\begin{align}
f_{score} = \hat{y}(h,r,t) = W_rh + r -W_rt \ ,
\end{align}
where $h$, $t$, and $r$ denote head, tail, and relation representations, respectively, and $W_r$ is a projection matrix that maps the head and tail entities into the relation space.
Then, a triplet loss function is calculated during the pre-training process to discriminate the positive and negative triplet pairs via a pairwise ranking loss as:
\begin{align}
 \mathcal{L}_ {pre} = \sum_{\forall(h,r,t) \in \mathcal{D}}&-{\ln\sigma\left(\hat{y}(h,r, \Bar{t} ) - \hat{y}(h,r,t)\right)} + \lambda \vert \vert  \theta \vert \vert _2 ^2 ,  
\end{align}
where $\mathcal{D}$ is the set of all triplets in the knowledge graph, $(h, r, \Bar{t})$ denotes the negative triplet, and $\theta$ is a $L_2$ regularization parameter.

After pre-training, we then fine-tune our model for the prediction task. 
Specifically, we calculate a coefficient score to estimate the relationship between each product and its status pair.
We first combine the two entities followed by a Multi-Layer Perceptron (MLP), as:
\begin{equation}
    \hat{y}_{h,t} = \hat{y} (h,t) = \langle \phi(h), \phi(t)\rangle \ = MLP \left (W_rh \lVert  W_r t \right ) ,
\end{equation}
where $h$ and $t$ denote the learned representations of pairs of cosmetics and their status, respectively, $\lVert $ is the concatenation operator, $W_r$ is the transformation matrix corresponding to the relation between $h$ and $t$, and $\phi(\cdot)$ represents the final vector embeddings computed from Eq. \ref{eq11}.
Then, a binary cross entropy loss function is used for training classification. 
The loss function for fine-tuning task is computed as:
\begin{align} 
 \mathcal{L}_{fn}  = 
-\sum_{
\forall(P_k, S_i) \in \mathcal{P}}  &  y_{P_k, S_i}\log\left(\hat{y}_{P_k, S_i}\right) \nonumber \\
& + ( 1 - y_{P_k, S_i})\log\left((1-\hat{y}_{P_k, S_i})\right) \ ,
\end{align}
where $P_k$ and $S_i$ are the cosmetic product and status entities, $y_{P_k, S_i}$ and $\hat{y}_{P_k, S_i}$ are the ground-truth and predicted outputs, and $\mathcal{P}$ is the set of the training cosmetic products and its status.


\section{Experiments}

In this section, we perform a series of comprehensive experiments to assess the effectiveness of our proposed model and compare it with existing state-of-the-art baselines. 
We first evaluate the model's performance on link prediction tasks. 
Then, we conduct ablation studies to investigate the impact of various relation types and the residual connections on the model's overall performance.

\subsection{Experimental Settings}


\subsubsection{ Evaluation Metrics}

Since our task is a binary classification problem, we utilized several evaluation metrics, including accuracy ($Acc$), precision ($P$), recall ($R$), and $F_1$ to evaluate the performance of our proposed model.
The evaluation metrics are defined as follows:
\begin{align}
    &Acc = \frac{|D^+_{true} \cup D^-_{true}|}{|D^+ \cup D^-|} \ , \ 
    P = \frac{|D^+_{true}|}{|D^+|} \ , \notag\\
    &R = \frac{|D^+_{true}|}{|D^+_{true} \cup D^-_{false}|} \ , \      
    F_1 = 2 \frac{ P R } {(P + R)} \ ,
\end{align}
where the $D^+_{true}$ and $D^-_{true}$ denote the correct predictions for positive and negative cosmetic products, respectively, and $D^{-}_{false}$ is the false predictions for negative cosmetics.  


\subsubsection{Baselines}

To verify the effectiveness, we compare our model to relevant translation-based methods and GNN models, which have gained remarkable success in knowledge graph representation learning.
Translation-based models learn representations by mapping the entities and their relations into latent space through translations.

\begin{itemize}
\item \textbf{TransE} \cite{bordes_2013}. 
TransE is a translational distance-based model for learning embeddings of entities and relations in knowledge graphs. 
The relationships between entities are modeled as translations in the embedding space. For a given triple $\langle h,r,t \rangle$, TransE aims to ensure that the embedding of the head entity plus the embedding of the relation is close to the embedding of the tail entity, i.e., $h+ r = t $.

\item \textbf{TransR} \cite{lin_2015}. 
In TransR, entities and relations are embedded in different vector spaces. 
While TransE represents relations as translations in a single embedding space, TransR introduces a specific relation space for each.
For a given triple $\langle h,r,t \rangle$, the model also learns a projection matrix $M_r$ for each relation $r$, mapping entity embeddings into the relation space, as: $M_r h + r = M_r t $. 

\end{itemize}
    
Furthermore, we also compare our model with recent GNNs, which could learn high-order sub-structures and semantic relations in KGs.
There are three GNN baselines, including KGNN, KGAT, and LiteralKG models, as:

\begin{itemize}
\item \textbf{KGNN} \cite{lin_2020_kgnn} extend traditional GNNs to capture higher-order relationships within the knowledge graph. 
The iterative propagation enables information from non-adjacent entities to influence the embedding of a target entity, enhancing its interaction prediction capability.

\item \textbf{KGAT} \cite{wang_2019} applies an attention mechanism to weigh the influence of each neighboring entity differently. 
This allows the model to prioritize more relevant connections, refining the embedding for each entity based on the most relations.
Moreover, KGAT can propagate information across multiple hops, capturing complex, multi-step relationships between entities. 

\item \textbf{LiteralKG} \cite{DBLP:journals/access/HoangNLLNL23} introduces an attentive propagation with residual connection and identity mapping to capture the complex relations between entities in the knowledge graph.
The model also could handle different types of entities' features, such as numerical and textual attributes, via a fusing mechanism.

\end{itemize}

\subsubsection{Implementation Details}

Our model was initially pre-trained with the self-supervised learning task without relying on the label information.
Thereafter, we fine-tuned the HaCKG model with the fine-tuning loss function for link prediction tasks.
For the fine-tuning process, we conducted experiments by randomly splitting the dataset into training, validation, and testing sets with proportions of 60\%, 20\%, and 20\%, respectively.

Our model implementation is based on the PyTorch library, which is available in the following open source repository\footnote{\url{https://github.com/NSLab-CUK/Halal-or-Not}}.
All experiments were conducted on two GPU servers, each equipped with four NVIDIA RTX A5000 GPUs (24GB RAM per GPU).
The Adam optimizer \cite{kingma_2014_adam} was utilized during both the pre-training and fine-tuning phases.
We used Leaky ReLU as the activation function.
We employ early stopping and optimize the hyperparameters using grid search.
The learning rate was selected from the range $ \{0.0001, {0.00005}, 0.00001\}$. 
The hidden dimensions of the GNN layers were chosen from $ \{16, 32, {64}, 128\}$, while the number of GNN layers was varied among $\{1, {2}, 3, 4, 5\}$.
For a fair comparison with baseline methods, we tuned the hyper-parameters within similar ranges, including learning rate, hidden dimensions, and the number of layers.

\subsection{ Results on Link Prediction}

\begin{table}[tb]
\centering
\caption{
The performance of our proposed model and baselines in terms of accuracy, recall, precision, and $F_1$. 
The top two are highlighted by \textbf{first} and \underline{second}.
}
\begin{adjustbox}{width=\linewidth}
\begin{tabular}{lcccc }
\toprule
\textbf{Model} 
& \textbf{Accuracy}	
& \textbf{Recall} 
& \textbf{Precision} 
& $F_1$
\\
\midrule

TransE \cite{bordes_2013}
&0.6358
&0.6125
&0.6072
&0.6098
\\
TransR \cite{lin_2015}
&0.6174
&0.6286
&0.5912
&0.6093
\\
\midrule
KGNN \cite{lin_2020_kgnn}
&0.7845 
&0.7361
&0.7948
&0.7643
\\
KGAT \cite{wang_2019}
&\underline{0.8713}
&0.8304	
&\underline{0.9043}
&0.8658
\\
LiteralKG \cite{DBLP:journals/access/HoangNLLNL23}
&0.8705
&\underline{0.8428}
&0.8922
&\underline{0.8668}
\\
\midrule
HaCKG (Ours)
&\textbf{0.9657}
&\textbf{0.9573}
&\textbf{0.9794}
&\textbf{0.9682}
\\
\bottomrule
\end{tabular}
\end{adjustbox}
\label{tab:node_classification}
\end{table}

Table \ref{tab:node_classification} shows the performance of our proposed model and baselines in terms of accuracy, recall, precision, and $F_1$.
We have the following observations:
(1) Our proposed model with pre-training outperformed baselines in all the evaluation metrics, showing the model's superiority compared to baselines.
For example, our proposed model reached the highest value at $0.9794$ regarding precision metric. 
This indicates that our model could learn the numerical information of entities well to maximize the relations between entities and capture complex entities' relations via the attention mechanism. 
(2) Translation-based models, i.e., TransE and TransR, showed low performance in predicting halal or haram status.
We argue that these models only capture the local relations independently, e.g., cosmetics and ingredients relations or ingredients and their properties, which could overlook high-order relations between entities, resulting in poor performance. 
(3) Several recent GNN-based models, i.e., KGAT and LiteralKG, have shown competitive performance compared to our model.
This indicates that learning relational information between entities is an essential factor that could contribute to the overall performance.
The superior performance of the model verifies that our model can capture well structural relations between cosmetics and ingredients and can help improve cosmetic representations.

\subsection{Model Analysis}



\subsubsection{Performance According to the Number of Layers}


\begin{figure}[tb]
\centering 
\includegraphics[width= 0.9 \linewidth]{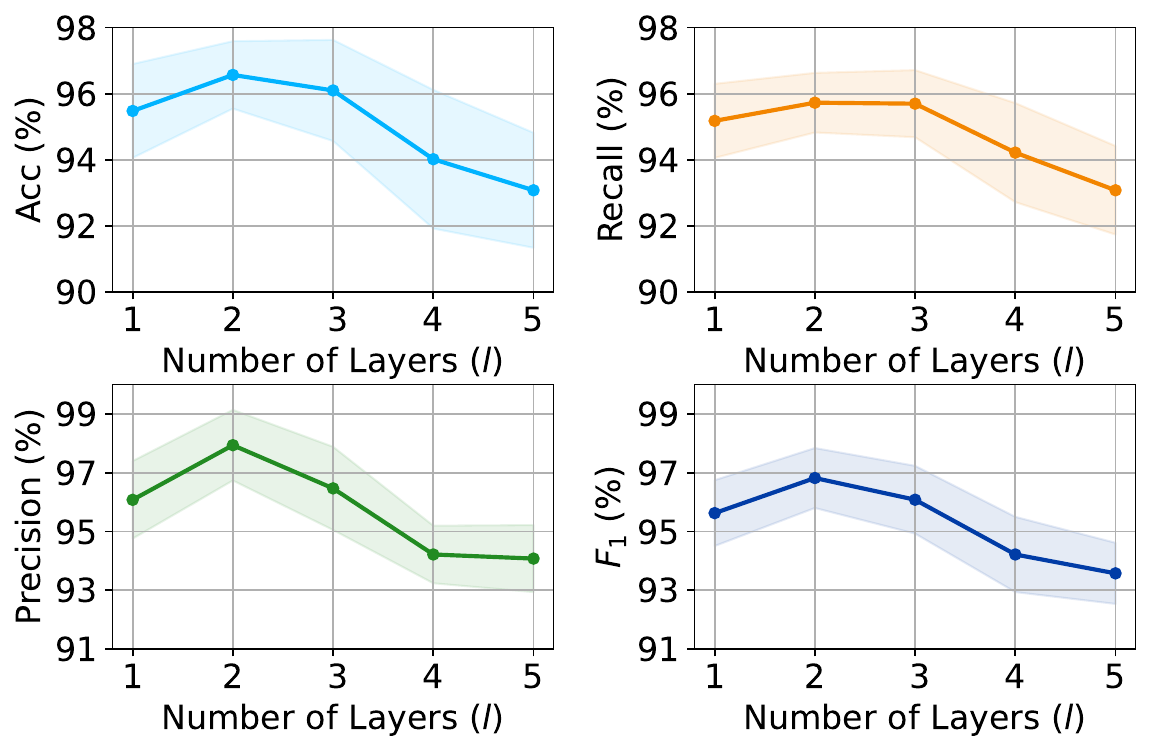}
\caption{The model accuracy according to the number of layers.
}  
\label{fig:Num_layers}
\end{figure}

We further analyze the impact of the number of layers in our proposed model on different metrics, as shown in Figure \ref{fig:Num_layers}.
We observed that: 
(1) The model's performance improved as $l$ increased, reaching its peak at $l=2$ and $l=3$.
This is because each entity aggregates information from its neighbors and those neighbors’ neighbors (2-hop).
That is, the proposed model captured sufficient local structures, enabling the model to learn robust representations for each entity.
(2) When the number of layers increased, e.g., $l >3$, the model performance was decreased as entities could receive redundant and overly smoothed information from distant nodes.
We argue that this could be the over-smoothing problem, which limits the model’s ability to distinguish between different graph structures, reducing model accuracy.

\subsubsection{Effect of Pre-training, Numerical Properties, and Residual Connection}

\begin{table}[t]
\centering
\setlength{\tabcolsep}{5 pt}
\caption{The effect of pre-training, numerical properties, and residual connection. The top two are highlighted by \textbf{first} and \underline{second}.}
\begin{adjustbox}{width=\linewidth}
\begin{tabular}{lcccc }
\toprule
\textbf{Model} & \textbf{Accuracy}	
& \textbf{Recall} 
& \textbf{Precision} 
& {$F_1$}
\\
\midrule

w/o pre-training  
&0.9171 
&0.8804 
&0.9043 
&0.8921

\\
w/o numerical attributes 
&\underline{0.9471}
&\underline{0.9136}
&\underline{0.9555}	
&\underline{0.9341}
\\
w/o residual connection 
&0.9108 
&0.9036 
&0.8958 
&0.8996
\\
HaCKG (Ours) 
&\textbf{0.9657}
&\textbf{0.9573}
&\textbf{0.9794}
&\textbf{0.9682}
\\ 
\bottomrule
\end{tabular}
\end{adjustbox}
\label{tab:abstudy}
\end{table}
To verify the impact of pre-training strategies, the use of numerical properties, and the residual connection, we conduct an ablation study by considering four variants of our proposed model, as shown in Table \ref{tab:abstudy}.
Specifically, for the first variant, i.e., w/o pre-training, we disable the pre-training step and directly train the model by using the loss function for prediction.
For the second variant, i.e., w/o numerical properties, we remove the numerical features of ingredient properties and replace them with random attributes.
Last, we disable the residual connection in our model to test the model performance.
We observed that:
(1) Removing the pre-training process and residual connection degrades the model performance.
Without pre-training, the proposed model consistently underperforms the overall model. 
This is because the model fails to learn the explicit structural information between triplets, resulting in the granularity of triplets and poor performance.
(2) Without using numerical properties, the model performance was slightly reduced, e.g., from 0.9657 to 0.9471 in terms of accuracy measurement.
We argue that ingredient properties, e.g., toxic scores and restriction state of ingredients, could contribute to the model to capture distinguished features between ingredients and products.
(3) The use of residual connections contributes significant improvements to the overall model performance.
Although residual connections have additional costs when stacking more GNN layers, we argue that our proposed model with residual connections is the key success to significantly improving model performance.
This is because the pre-trained model could learn the structural relations between cosmetics to distinguish various types of cosmetics in the self-supervised setting, capturing the natural relationship between cosmetics.
To sum up, using pre-training, numerical properties, and residual connections could contribute to the model's performance.

\section{Conclusion and future work}

In this work, we propose HaCKG, a pre-trained relational graph attention network model, to learn the halal status of cosmetic products.
We first construct a cosmetic knowledge graph that represents the natural relations between products, ingredients, and their properties.
The model is then pre-trained to learn the structural similarity between triplets in the cosmetic knowledge graph, followed by a prediction task.
By doing so, the model could learn the knowledge graph structures and relations between entities in the knowledge graph, which aligns with the nature of the knowledge graph.
Our extensive experiments demonstrate that the proposed model consistently outperforms existing baselines.
While the current pre-training strategy focuses on learning structural similarities between triplets, there are several future directions for further exploration.
For example, exploring advanced pre-training tasks, such as contrastive learning, could further enhance the model's understanding of the knowledge graph.

\section*{Acknowledgment}

This work was supported 
in part by the National Research Foundation of Korea (NRF) grant funded by the Korea government (MSIT) (No. 2022R1F1A1065516 and No. 2022K1A3A1A79089461) (O.-J.L.),
in part by the “Leaders in INdustry-university Cooperation 3.0” Project funded by the Ministry of Education and National Research Foundation of Korea (O.-J.L.), 
%
and
in part by the R\&D project “Development of a Next-Generation Data Assimilation System by the Korea Institute of Atmospheric Prediction System (KIAPS)” funded by the Korea Meteorological Administration (KMA2020-02211) (H.-J.J.).

\bibliography{aaai25}

\end{document}